\documentclass[lettersize,journal]{IEEEtran}
\usepackage{amsmath,amsfonts}
\usepackage{algorithmic}
\usepackage{algorithm}
\usepackage{array}
\usepackage{textcomp}
\usepackage{stfloats}
\usepackage{url}
\usepackage{verbatim}
\usepackage{graphicx}
\usepackage{cite}
\usepackage{booktabs}
\usepackage{multirow}
\usepackage[normalem]{ulem}
\usepackage{subcaption}
\useunder{\uline}{\ul}{}

\begin{document}

\title{Leveraging Auxiliary Task Relevance for Enhanced Bearing Fault Diagnosis through Curriculum Meta-learning}

\author{Jinze Wang,~\IEEEmembership{Student Member,~IEEE,} Jiong Jin$^\dag$,~\IEEEmembership{Member,~IEEE,} Tiehua Zhang$^\dag$,~\IEEEmembership{Member,~IEEE}, Boon Xian Chai,   Adriano Di Pietro, and Dimitrios Georgakopoulos,~\IEEEmembership{Senior Member,~IEEE}
        % <-this % stops a space
\thanks{Jinze Wang, Boon Xian Chai, Adriano Di Pietro are with Aerostructures Innovation Research Hub (AIRHub), Swinburne University of Technology, Melbourne, Australia (email: \{jinzewang, bchai, adipietro\}@swin.edu.au).}
\thanks{ Jiong Jin, Dimitrios Georgakopoulos are with the School of Science, Computing and Engineering Technologies, Swinburne University of Technology, Melbourne, Australia (email: \{jiongjin, dgeorgakopoulos\}@swin.edu.au).}
\thanks{Tiehua Zhang is with the Department of Computer Science and Technology, Tongji University, Shanghai, China (email: tiehuaz@tongji.edu.cn).}
\thanks{$^\dag$ Corresponding author: Jiong Jin (jiongjin@swin.edu.au), Tiehua Zhang (tiehuaz@tongji.edu.cn).}}
% <-this % stops a space
%\thanks{Manuscript received XXX, 2025}


% % column for its text to clear the IEEEpubid mark.

\maketitle

\begin{abstract}
The accurate diagnosis of machine breakdowns is crucial for maintaining operational safety in smart manufacturing. Despite the promise shown by deep learning in automating fault identification, the scarcity of labeled training data, particularly for equipment failure instances, poses a significant challenge. This limitation hampers the development of robust classification models. Existing methods like model-agnostic meta-learning (MAML) do not adequately address variable working conditions, affecting knowledge transfer. To address these challenges, a Related Task Aware Curriculum Meta-learning (RT-ACM) enhanced fault diagnosis framework is proposed in this paper, inspired by human cognitive learning processes. RT-ACM improves training by considering the relevance of auxiliary sensor working conditions, adhering to the principle of ``paying more attention to more relevant knowledge", and focusing on ``easier first, harder later" curriculum sampling. This approach aids the meta-learner in achieving a superior convergence state. Extensive experiments on two real-world datasets demonstrate the superiority of RT-ACM framework.
\end{abstract}

\begin{IEEEkeywords}
 Fault diagnosis, meta-learning, few-shot, curriculum, task relevance.
\end{IEEEkeywords}

\section{Introduction}
\label{sec:introduction}
\IEEEPARstart{F}{ault} diagnosis is crucial for ensuring the safe and efficient operation of machines~\cite{wu2020few, meng2022intelligent}. Recently, methods based on deep learning (DL) have demonstrated considerable promise in accurately diagnosing machine faults through sensor signals~\cite{pang2020investigation,hasan2023wasserstein}. An advanced ensemble convolutional neural network (CNN) has been introduced for this purpose, utilizing diverse local minima to enhance model adaptability~\cite{wen2022new}. Furthermore, a capsule network that builds upon CNNs has been proposed for fault classification in bearings~\cite{zhu2019convolutional}. This network combines an inception block and a regression branch, thereby increasing the model's versatility. However, a notable challenge faced by DL methods is the requirement for extensive training data. Accumulating such data in real-world scenarios is often not feasible due to associated costs and logistical challenges~\cite{chen2020robust}. When machines exhibit faults, they are typically partially (e.g., showed down) or fully (e.g., turned off) deactivated for safety reasons, resulting in a scarcity and variety of fault data~\cite{hu2021task}. Although intentional fault inductions on lab-based machines might provide data, such interventions in operational settings are strictly prohibited. This situation intensifies the challenge of developing reliable DL classifiers for fault diagnosis in real-world contexts~\cite{hu2018imbalance}.

Three key strategies have emerged to tackle the challenge of data scarcity in creating reliable fault diagnosis models: Data Augmentation (DA)~\cite{li2020intelligent}, Transfer Learning (TL)~\cite{zhuang2020comprehensive}, and Few-Shot Learning (FSL) approaches~\cite{chen2022meta}. DA aims to expand and diversify datasets by producing additional data from the limited original samples available. For example, data enhancement techniques that incorporate Gaussian noise, masking noise, and signal translation have been explored in~\cite{li2020intelligent, zhao2024multi}. Moreover, generative models, including variational autoencoders and generative adversarial networks, have been employed to generate sensor fault data for training DL-based diagnostic models in~\cite{ gao2020data}. However, a notable limitation of DA is the potential degradation in data quality and precision, especially when original datasets are sparse. This might reduce the effectiveness of models trained on such data.

To circumvent the limitations of DA, the use of prior knowledge from related tasks has been explored, leading to the concept of Transfer Learning (TL). TL methods aim to transfer knowledge gained from one task to improve performance on another closely related task. For instance, a comprehensive one-dimensional CNN architecture, supplemented with fine-tuning mechanisms specifically for fault diagnosis, has been presented\cite{wu2020few, zhu2019new}. Similarly, a CNN-based model for diagnosing faults in bearings has been proposed, with its basis being knowledge garnered from laboratory bearings data\cite{yang2019intelligent}. However, a potential drawback of TL techniques is their inherent emphasis on adapting models to target tasks, which may hinder their broad applicability, especially in environments with limited data~\cite{chang2022meta}. Due to this constraint, there exists a growing need to enhance the overall generalization capabilities of TL methods for more effective fault diagnosis in situations characterized by data sparsity.

Few-shot learning (FSL) stands out for its capability to generalize across tasks, not merely adapting to one. It achieves this by leveraging insights from various auxiliary tasks, a method commonly known as meta-learning~\cite{naik1992meta}. This stands in contrast to transfer learning, where the primary focus is adaptation to a distinct task. FSL delves deeper by leveraging insights from associated tasks to benefit the primary FSL task~\cite{wang2020generalizing}. In practical settings, machines typically operate under varied conditions, gathering substantial sensor data for each specific condition is both expensive and, in some cases, impractical~\cite{wang2024few}. Yet, when data from varied operational conditions are utilized, a multitude of auxiliary tasks can be formulated. This inherent flexibility renders meta-learning especially fitting and potent for few-shot fault diagnosis scenarios. Such FSL models specifically designed for fault diagnosis have been proposed in~\cite{long2023customized,vu2024few}. These models frequently incorporate metric-based meta-learning methods. 

\begin{figure}[t]
  \centering
  \includegraphics[width=.7\linewidth]{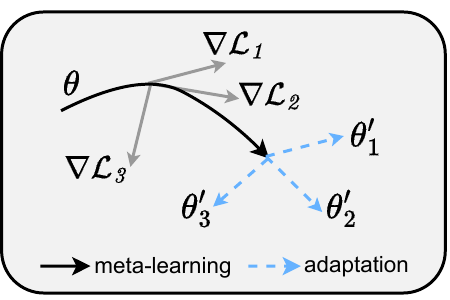}
  \caption{Learning process of MAML, which optimizes for a representation $\theta$ that can quickly adapt to new tasks.}
  \label{fig:maml}
\end{figure}

Metric-based meta-learning methods have demonstrated potential in feature extraction; however, their effectiveness is often limited by data sparsity and variations in auxiliary task distributions. Model-Agnostic Meta-Learning (MAML)~\cite{finn2017model}, an optimization-based meta-learning framework, offers a different approach by focusing on developing models with strong generalization capabilities. MAML seeks to identify a set of initial parameters that can be fine-tuned with minimal data to achieve optimal performance across a series of new tasks. As shown in Fig.~\ref{fig:maml}, MAML optimizes the model to establish a representation, $\theta$, that facilitates rapid adaptation to new tasks. By leveraging a small amount of target task data for fine-tuning, MAML achieves notable classification accuracy in few-shot bearing fault diagnosis. For instance, Zhang et al.~\cite{zhang2021few} proposed a MAML-based framework employing an enhanced meta-relation network for bearing fault diagnosis with limited data. Zhou et al.~\cite{zhou2024prior} further advanced this concept by integrating prior knowledge into MAML, enabling more effective utilization of prior knowledge in both training and testing phases. Chen et al.~\cite{chen2022meta} developed a MAML-based model featuring a four-layer CNN designed for feature extraction and rapid adaptation, allowing efficient fault diagnosis under diverse working conditions. Additionally, Cui et al.~\cite{ren2023few} employed a federated meta-learning approach combining representation encoding with MAML to enhance performance in distributed fault diagnosis tasks. Although these studies have achieved some performance improvements and emphasized that MAML learns to learn, they have overlooked the difficulty level of each task and have not considered the human learning experience of easy-to-hard curriculum learning. A specialized task-sequencing strategy, task-sequencing meta-learning (TSML)~\cite{hu2021task}, has been introduced for few-shot fault diagnosis. This approach arranges auxiliary tasks from simple to complex, aiming to enhance knowledge adaptability during meta-learning. However, it relies on k-means clustering to evaluate task difficulty, which overlooks the intricate relationships within temporal data, limiting its effectiveness to simpler datasets. Furthermore, these existing studies have not considered the relevance between different tasks and the target task, as unrelated auxiliary tasks can adversely affect the target task.

To simultaneously consider data sparsity and task diversity challenges in fault diagnosis, we propose a novel Related Task Aware Curriculum Meta-learning (RT-ACM) framework that integrates human-inspired easy-to-hard curriculum learning with task relevance considerations. RT-ACM leverages the meta-learning paradigm to capture the relationships among patterns from diverse tasks, enabling more effective knowledge transfer under limited data conditions. Building on the foundation of MAML, the framework is designed to enhance adaptation to target tasks with minimal training data by treating fault diagnosis tasks under varying working conditions as individual tasks. The meta-training process incorporates two key components: a related task-aware meta-learning mechanism that emphasizes task relevance, and a curriculum strategy that systematically addresses task diversity, facilitating improved learning efficiency and generalization.

To enhance the meta-learner, we exploit the heuristic idea of "paying more attention to more relevant knowledge," adopting a related task-aware meta-learning approach. This strategy encourages the model to prioritize learning from tasks that are highly relevant, thereby improving generalization. Each meta-training round is structured into two stages: first, the meta-learner is updated using a MAML-like process; second, a new batch of tasks is conditionally re-sampled, focusing on the most challenging working conditions as determined by validation scores. Specifically, we propose a curriculum strategy to refine the task sampling process during MAML training. Guided by the principle of an "easier-first, harder-later" paradigm, this approach directs the meta-learner toward an optimized state. A teacher recommender is pre-trained for each auxiliary task, leveraging the best validation score to assess task difficulty and dynamically designing the task sampling curriculum at each meta-training step.

Overall, our main contributions can be summarized in three key aspects:

\begin{enumerate}
  \item To the best of our knowledge, this is one of the early studies that examines the extent to which knowledge can be transferred from auxiliary tasks to the target task by differentiating the relevance of working conditions. A novel related task-aware curriculum meta-learning (RT-ACM) enhanced fault diagnosis framework is proposed, which leverages both transferred knowledge and the diversity of tasks within the target task to mitigate the issue of data limitation.
  \item A meta-training process is formulated, which incorporates related task-aware meta-learning and a curriculum strategy to leverage auxiliary tasks while balancing the high diversity among different tasks. The related task-aware meta-learning follows the principle of ``paying more attention to more relevant tasks", while the curriculum strategy follows the principle of ``easy first, hard later." Moreover, we design a teacher recommender to evaluate task difficulty in temporal data.
  \item Extensive experiments have been conducted on two real-world datasets to validate the superiority of RT-ACM against state-of-the-art approaches. Furthermore, we have made our source code publicly available to contribute further to advancements in this field.
\end{enumerate}

The remainder of this article is structured as follows: Section \ref{sec:proposedmethod} provides a detailed description of the proposed method. Section \ref{sec:casestudy} presents a series of case studies to evaluate the effectiveness of RT-ACM. Finally, Section \ref{sec:conclusion} offers conclusions and outlines directions for future work.

\section{Proposed RT-ACM}
\label{sec:proposedmethod}
Let $A$ auxiliary tasks be represented as $\left\{ T^\mu \right\}_{\mu=1}^A$, where each auxiliary task $T^\mu$ comprises a dataset $\left\{(x^\mu_i, y^\mu_i)\right\}_{i=1}^M$. Here, $x^\mu_i \in \mathcal{R}^{1 \times D}$ corresponds to the $i$th input sample derived from 1-D vibration sensor signals, a common data source in this domain. These signals are segmented using a sliding window of length $D$ to produce samples, and $y^\mu_i$ denotes the sensor health status label associated with $x^\mu_i$. The total number of samples in $T^\mu$ is given by $M$. Similarly, a target task is defined as $T^{tar} = \left\{(x^{tar}_i, y^{tar}_i)\right\}_{i=1}^{M^{tar}}$, where the number of available training samples, $M^{tar}$, is limited, reflecting the scarcity of labeled data. Each $x^{tar}_i \in \mathcal{R}^{1 \times D}$ represents the input data for the $i$th sample, while $y^{tar}_i$ is its corresponding label.

\begin{figure*}[t]
\centering
\includegraphics[width=1\textwidth]{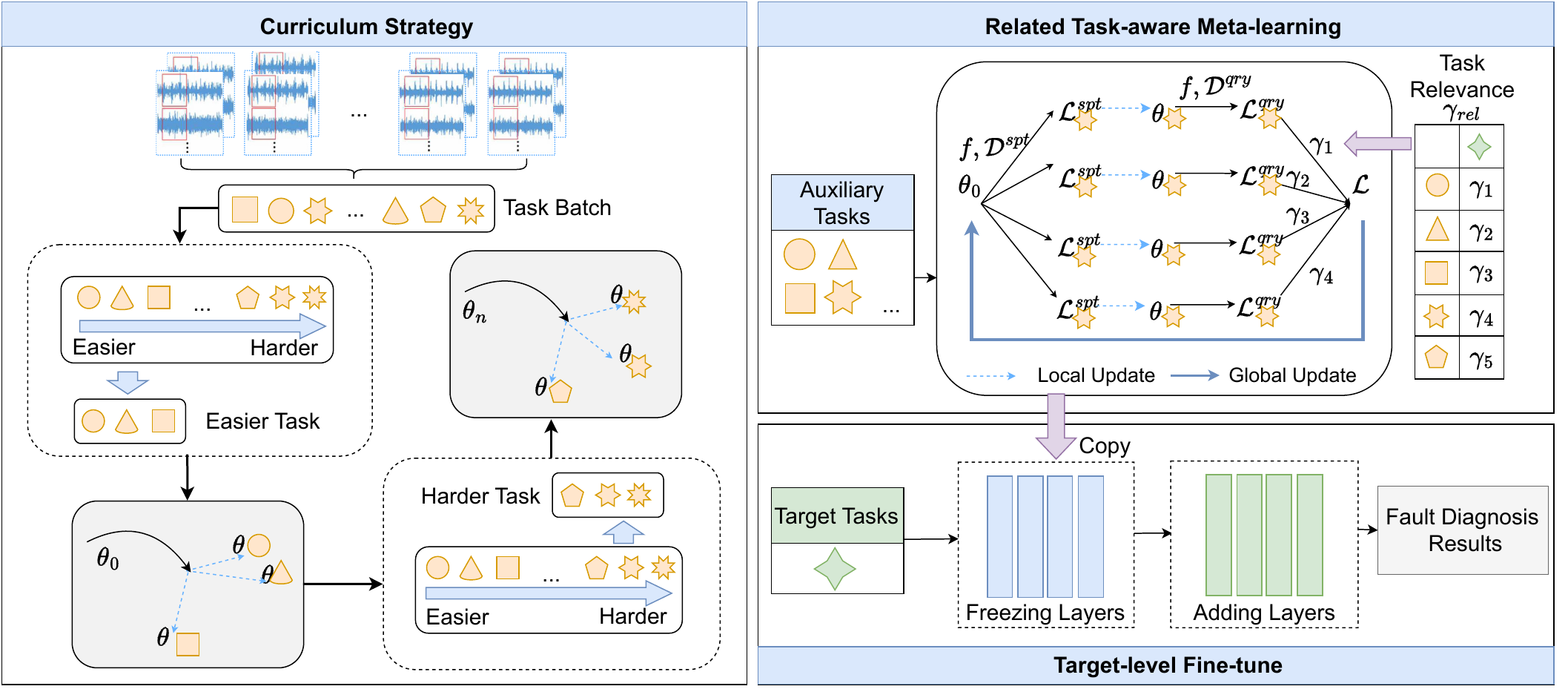}
\caption{The overall framework of our proposed RT-ACM.}
\label{fig:framework}
\end{figure*}

The structure of RT-ACM is depicted in Fig.~\ref{fig:framework}, consisting primarily of two components: meta-level training on auxiliary tasks and fine-tuning on the target task. Specifically, the meta-level training employs meta-learning to extract shared sensor vibration patterns, adhering to the principles of ``paying more attention on more relevant knowledge" and ``easy first, hard later," as determined by the teacher recommender. Meanwhile, the target-level fine-tuning aims to capture precise vibration patterns within the target task to enhance prediction accuracy.

\subsection{Meta-level Training on Auxiliary Tasks}
To distill knowledge from auxiliary tasks and leverage vibration patterns, we enhance the MAML framework by incorporating long short-term memory (LSTM)~\cite{hochreiter1997long} for meta-learning updates. Notably, we propose a related task-aware meta-learning approach combined with a curriculum strategy, enabling knowledge transfer by evaluating the relevance of patterns across tasks.

\subsubsection{Basic Theory of MAML}
As a subset of meta-learning, the MAML algorithm differs from traditional deep learning by focusing on training a model's initialization parameters within its outer loop. By optimizing these initialization parameters to be broadly applicable, the model can adapt to new tasks with limited data efficiently, requiring only a few gradient updates. From a learning perspective, MAML aims to uncover general internal representations that promote effective transferability across diverse tasks. As a result, minimal parameter adjustments are needed when tackling new tasks.

Following~\cite{chang2022meta}, fault diagnosis across various working conditions can be framed as a single task within the meta-learning framework. Accordingly, the vibration signals from auxiliary tasks $\{ T^\mu \}{\mu=1}^A$ are represented as $\mathbb{D}_{meta}^{(aux)}$, while the vibration signals from target tasks $T^{tar}$ are partitioned into training sets $\mathbb{D}_{train}^{(tar)}$ and testing sets $\mathbb{D}_{test}^{(tar)}$. Each working condition $y_m$ is treated as an individual meta-learning task, with a support set $\mathcal{D}^{spt}_{y_m}$ for training and a query set $\mathcal{D}^{qry}_{y_m}$ for evaluation. Ultimately, the objective is to utilize the auxiliary tasks to train a meta-learner $F_w$ that encapsulates knowledge across various working conditions.

Specifically, in each iteration of MAML, the process involves a local update and a global update on a batch of sampled tasks. During the local update phase, the model parameters $\theta$ are adjusted on the support set $\mathcal{D}^{spt}$ for each task. Subsequently, in the global update phase, $\theta$ is updated using gradient descent to minimize the aggregate loss computed on the query set $\mathcal{D}^{qry}$ across all tasks.

\begin{itemize}
    \item {Local Update:} A batch of tasks is first sampled from $\mathbb{D}^{(aux)}_{meta}$. For each selected task, a subset of vibration signals is randomly drawn from its support set $\mathcal{D}^{spt}_{y_m}$ and query set $\mathcal{D}^{qry}_{y_m}$. The training loss is then computed on $\mathcal{D}^{spt}_{y_m}$, and the parameters $\theta$ are locally updated with a single gradient descent step:

    \begin{equation}
    \label{equ2}
     \theta_{y_m}^{\prime}=\theta-\alpha \nabla_{\theta} \mathcal{L}_{y_m} (f_\theta,\mathcal{D}^{spt}_{y_m}),
    \end{equation}
    where $\mathcal{L}$ is the cross-entropy loss; $\alpha$ is the local learning rate; and $\theta_{y_m}^{\prime}$ is the locally updated recommender parameters on each task.

    \item {Global Update:} The testing loss is evaluated on each $\mathcal{D}^{qry}_{y_m}$ using the corresponding locally updated parameters $ \theta_{y_m}^{\prime}$. The goal of the MAML global update is to update the initialization $\theta$ by performing a single gradient descent step on the aggregated testing losses across all tasks:
    
    \begin{equation}
    \label{equ3}
     \theta =\theta -\beta \nabla_{\theta} \sum \mathcal{L}_{y_m} (f_{\theta'_{y_m}},\mathcal{D}^{qry}_{y_m}),
    \end{equation}
    where $\beta$ is the global learning rate. 
\end{itemize}

The optimal parameters for the meta-learning model are achieved through iterative optimization within the inner (local update) and outer (global update) loops. Consequently, the objective of meta-optimization is to minimize the task-specific loss function:

\begin{equation}
\label{equ}
\min_\theta \sum \mathcal{L}_{y_m}(f_{\theta'_{y_m}}) = \sum \mathcal{L}_{y_m}(f_\theta -\alpha\nabla_\theta\mathcal{L}_{y_m}(f_\theta )).
\end{equation}

\subsubsection{Related Task-aware Meta-learning}
{Directly transferring patterns from auxiliary tasks to the target task can introduce noise, potentially degrading model performance. To address this, we adhere to the principle of ``paying more attention on more relevant knowledge" and incorporate the \textit{relevance of working conditions} across tasks during the global update. By calculating task relevance (e.g., $\gamma_{rel}$) based on variations in vibration signal patterns, we adapt gradients across tasks with attention to their relevance.

To be specific, we employed autoencoders, a class of unsupervised neural networks, to ascertain the relevance among diverse auxiliary tasks with target task. For three distinct auxiliary tasks as example, represented by datasets $ \mathcal{Y}_0$, $\mathcal{Y}_1$, and $\mathcal{Y}_2$, and a target task $\mathcal{Y}_{tar} $, an autoencoder was trained on their amalgamated data. This model comprises two parts: the encoder $f_{\text{enc}}$ and the decoder $f_{\text{dec}}$. In essence, the encoder compresses the input $x$ into a latent representation $y$, and the decoder attempts to reconstruct $x$ from $y$. Post-training, the encoder transforms each dataset into this compact latent space, yielding representations $\mu$. For a holistic view of each task, we calculate the mean vector of these representations, denoted as $\gamma_{rel}$. The relevance between auxiliary tasks and target task is then quantified using the Euclidean distance:

\begin{equation}
\label{equ4}
\gamma_{rel_{i,t}} = \frac{1}{\sqrt{1+\sum_{k} (\mu_{i,k} - \mu_{t,k})^2}}
\end{equation}
where indices $i$ designate different auxiliary tasks, $t$ is the target task and $k$ traverses the latent dimensions. The derived distances, namely $\gamma_{rel_{0,t}}$, $\gamma_{rel_{1,t}}$, and $\gamma_{rel_{2,t}}$, serve as task relevance, with values closer to 1 suggesting more related tasks.} In essence, if an auxiliary task exhibits greater relevance to the target task, the gradient is adjusted to facilitate faster updates in that direction. Consequently, Eq.~(\ref{equ2}) is revised as follows: 

\begin{equation}
\label{equ5}
 \theta_{y_m}^{\prime}=\theta-\alpha \nabla_{\theta} [\mathcal{L}_{y_m} (f_\theta,\mathcal{D}^{spt}_{y_m}) \times\gamma_{rel}].
\end{equation}

\subsubsection{Curriculum Strategy}
{To address the challenges posed by high task diversity, we propose a novel approach inspired by the curriculum learning paradigm to improve both the convergence speed and the generalization ability of the meta-learner. This approach adopts a sequential task presentation strategy, following the principle of ``easier tasks first, harder tasks later." To implement this, we introduce a \textit{teacher recommender}, which undergoes initial training for each auxiliary task and utilizes the highest validation score to estimate task difficulty. This difficulty metric guides the curriculum design for task sampling during each meta-training iteration. Specifically, to measure task difficulty, we split the samples into a training set $D_{\text{train}}$ and a validation set $D_{\text{valid}}$. A base LSTM model is trained extensively on $D_{\text{train}}$ to serve as the teacher recommender. The difficulty of a task is quantified using a performance metric $\Phi$ (e.g., AUC) evaluated on $D_{\text{valid}}$. The task difficulty, denoted as $\delta_c$, is inversely proportional to the highest validation score obtained across all model parameters $\theta$:

\begin{equation}
\label{equ6}
\delta_c \propto - \max_{\theta} \Phi( f_{\theta}, D_{\text{valid}_c} | D_{\text{train}_c} )
\end{equation}

Intuitively, a lower peak validation score indicates more challenging vibration signal pattern within a task.}

\begin{algorithm}[t]
	\renewcommand{\algorithmicrequire}{\textbf{Require:}}
	\caption{Related Task Aware Curriculum Meta-learning (RT-ACM) Enhanced Fault Diagnosis framework}
	\label{alg:1}
	\begin{algorithmic}[1]
		\REQUIRE $\mathbb{D}_{meta}^{(aux)}$; learning rates $\alpha$, $\beta$; number of shots $N$; max step of iterations $n$;
		\quad 
		\STATE Randomly initialize parameters $\theta$;
        \STATE Calculate the relevance between tasks by Eq.(\ref{equ4});
		\STATE Calculate the difficulty of auxiliary tasks by Eq.(\ref{equ6});
		\WHILE {not done}
		    \FOR{ $all$ $\mathbb{D}_i \in \mathbb{D}_{meta}^{(aux)}$}
		        \STATE Sample $N$ vibration signal from $\mathbb{D}_i$ as the adapt\_batch based on the difficulty; 
		        \STATE Evaluate: $\nabla_{\theta}\mathcal{L}_{y_m}(f_\theta,\mathcal{D}^{spt}_{y_m})$ using adapt\_batch;
		        \STATE Calculate the gradient update of $\theta'_{y_m}$ by Eq.(\ref{equ5});
		        \STATE Sample another $N$ signal from $\mathbb{D}_i$ as the eval\_batch based on the difficulty;
		    \ENDFOR
            \STATE Update $\theta$ using eval\_batch by Eq.(\ref{equ3});
         \ENDWHILE
         
        \STATE Freeze the first $l$ layers as new LSTM model, i.e., LSTM$_{frozen}$; 
        \STATE Fine-tune LSTM$_{frozen}$ using only the target data $\mathbb{D}_{train}^{(tar)}$;
        \STATE Predict next possible fault via Eq.(\ref{equ8});
        \STATE Calculate the prediction loss for each record via Eq.(\ref{equ9});
	\end{algorithmic}  
\end{algorithm}

\subsection{Target-level Fine-tune}
After obtaining the general pattern of auxiliary tasks through the meta-learning paradigm, it is important to further fine-tune the learning model for the target task. By investigating the freezing of layers and then fine-tuning, the network generalizes better than one trained directly on the target dataset~\cite{yosinski2014transferable}. This operation not only incorporates information from the auxiliary tasks but also allows for better adaptation to the target task~\cite{wang2023meta}.

After extracting general patterns from auxiliary tasks via the meta-learning paradigm, fine-tuning the model for the target task becomes essential~\cite{yosinski2014transferable}. By investigating the freezing of layers and then fine-tuning, the network generalizes better than one trained directly on the target dataset. This approach not only integrates knowledge from auxiliary tasks but also facilitates better adaptation to the target task~\cite{wang2023meta}.

Assuming the meta-trained model (e.g., LSTM) consists of $L$ layers, we freeze the first $l$ layers $(1 \leq l \leq L)$, add $n$ new layers after the frozen ones, and fine-tune the parameters using target task data to maximize the reuse of general parameters. Consequently, the target task input $\mathbf{e}_{hist}^{(tar)}$ is passed through the recurrent layer to compute the hidden state $\boldsymbol{h^u_{t_k}}$ at time $t_k$, as expressed by:

\begin{equation} \label{equ7} \boldsymbol{h^u_{t_k}} = LSTM_{frozen}(\mathbf{e}_{hist}^{(tar)}). \end{equation}

By doing this, the freezing-layer operation facilitates the creation of a network that achieves a more effective balance between the parameters learned from auxiliary tasks and those optimized for the target task, following the final fine-tuning of the model.

The probability distribution on all working conditions is calculated by softmax function:
\begin{equation}
\label{equ8}
\boldsymbol{\hat{y}} = softmax(f(\mathbf{h}^u_{t_k}))
\end{equation}
where $f$ is a fully connected layer to transform $\boldsymbol{h^u_{t_k}}$ into a $|\mathcal{P}|$-dimensional vector, and $|\mathcal{P}|$ is the number of faulty class in the target task. Hence, the objective function is defined by:
\begin{equation}
\label{equ9}
\mathcal{J} =  -\sum_{i=1}^{|\mathcal{P}|} \mathbf{y}[i]\cdot log \space (\hat{\mathbf{y}}[i])
\end{equation}
where $\mathbf{y}$ is an one-hot embedding of the ground-truth fault. 
Algorithm~\ref{alg:1} summarizes the training process of RT-ACM, consisting of meta training (lines 4-12), freezing layers and model fine tuning (lines 13-14), as well as fault diagnosis (lines 15-16).

\section{Case Study}
\label{sec:casestudy}
% \usepackage{multirow}
\begin{table*}[tb]
\centering
\caption{COMPARISON OF DL-BASED AND META-LEARNING-BASED MODELS \\(*The best results are highlighted in bold; the runner up is underlined.)}
\label{tab:results_CWRU}
\renewcommand{\arraystretch}{1.1}
\begin{tabular}{ccccccccccccccccc}
\toprule
\hline
                                 & \multicolumn{8}{c}{\textbf{3-way}}                                                                                                    & \multicolumn{8}{c}{\textbf{6-way}}                                                                                                    \\ \cline{2-17} 
\multirow{2}{*}{\textbf{Models}} & \multicolumn{4}{c}{\textbf{1-shot}}                               & \multicolumn{4}{c}{\textbf{5-shot}}                               & \multicolumn{4}{c}{\textbf{1-shot}}                               & \multicolumn{4}{c}{\textbf{5-shot}}                               \\
                                 & T1             & T2             & T3             & T4             & T1             & T2             & T3             & T4             & T1             & T2             & T3             & T4             & T1             & T2             & T3             & T4             \\ \hline
Cap-Net                          & 41.23          & 35.56          & 45.24          & 47.89          & 69.53          & 55.43          & 65.31          & 63.74          & 37.53          & 31.83          & 41.02          & 44.65          & 63.18          & 50.29          & 59.55          & 57.38          \\
WDCNN                            & 41.50          & 34.48          & 45.90          & 46.07          & 69.84          & 55.75          & 66.93          & 63.23          & 37.82          & 30.92          & 42.45          & 44.03          & 63.68          & 50.53          & 60.29          & 57.05          \\
MAML                             & 85.23          & 80.92          & 83.95          & 85.80          & 97.71          & 95.72          & 97.03          & 96.75          & 70.25          & 66.38          & 69.22          & 71.45          & 73.59          & 70.87          & 72.13          & 71.68          \\
TMSL                             & \uline{97.21}    & \uline{96.90}    & \uline{97.87}    & \uline{97.80}     & \uline{98.20}    & \uline{97.50}          & \uline{99.47}    & \uline{99.66}    & \uline{86.00}    & \uline{82.18}          & \uline{88.71}          & \uline{89.32}    & \uline{89.96}    & \uline{87.29}          & \uline{92.13}          & \uline{91.45}    \\
\textbf{RT-ACM}                   & \textbf{98.51} & \textbf{97.44} & \textbf{98.62} & \textbf{98.53} & \textbf{99.68} & \textbf{99.53} & \textbf{99.73} & \textbf{99.56} & \textbf{93.22} & \textbf{92.49} & \textbf{94.38} & \textbf{95.14} & \textbf{97.25} & \textbf{96.59} & \textbf{97.48} & \textbf{97.05} \\ \hline \bottomrule
\end{tabular}
\end{table*}

Rolling bearings are critical components in electric machines and are widely utilized in various mechanical systems. Due to prolonged rotational stress and exposure to high temperatures, bearing failures frequently occur during operation. As a result, research on fault diagnosis for rolling bearings is of paramount importance. This study concentrates on detecting faults in rolling bearings within electric motors, taking into account diverse working conditions and the limited availability of samples. To assess the performance of the proposed model, a series of case studies are carried out using two well-known datasets: the Case Western Reserve University (CWRU) Bearing Dataset~\cite{smith2015rolling} and the Paderborn University Rolling Bearing Dataset~\cite{lessmeier2016condition}.

Additionally, several state-of-the-art models are employed for comparison, including the capsule network (Cap-Net)~\cite{zhu2019convolutional}, wide deep CNN (WDCNN)~\cite{zhang2017new}, MAML, TMSL, CHAML, and PKAML. Among these, Cap-Net and WDCNN are deep learning-based models, while MAML, CHAML, and PKAML belong to the category of meta-learning-based approaches.

\begin{table}[t]
\centering
\caption{INFORMATION OF USED BEARINGS FROM CWRU DATASET}
\label{tab:info_cwru}
\begin{tabular}{ccc}
\toprule
\toprule
\textbf{Class Label} & \textbf{Fault Type} & \textbf{Fault Diameter (Inches)} \\
\midrule
1 & Healthy & 0 \\
\midrule
2 & Ball & 0.007 \\
3 & Ball & 0.014 \\
4 & Ball & 0.021 \\
\midrule
5 & Inner Race & 0.007 \\
6 & Inner Race & 0.014 \\
7 & Inner Race & 0.021 \\
\midrule
8 & Outer Race & 0.007 \\
9 & Outer Race & 0.014 \\
10 & Outer Race & 0.021 \\
\bottomrule
\bottomrule
\end{tabular}
\end{table}

\begin{table}[t]
\centering
\caption{DETAILED DESCRIPTIONS OF EACH TASK UNDER CWRU DATASET}
\label{tab:cwru_tasks}
\begin{tabular}{cccc}
\toprule
\toprule
\textbf{Task} & \textbf{Load (hp)} & \textbf{Number of Class} & \textbf{Motor Speed (r/min)} \\
\midrule
$T^1$   & 0         & 10              & 1797                       \\
$T^2$   & 1         & 10              & 1772                       \\
$T^3$   & 2         & 10              & 1750                       \\
$T^4$  & 3         & 10              & 1730                      \\
\bottomrule
\bottomrule
\end{tabular}
\end{table}

\subsection{Case A: CWRU Bearing Dataset}
\subsubsection{Dataset Description} 
The CWRU dataset is widely utilized to evaluate fault diagnosis methods~\cite{meng2022intelligent,zhu2019new,yang2023feature}. The selected rolling bearing dataset from CWRU includes a comprehensive set of fault types, as shown in Table \ref{tab:info_cwru}. It covers four distinct electric machine health states, including one normal state and three fault states. The fault sizes are specified by their diameters, which measure 0.007, 0.014, and 0.021 inches. As outlined in Section~\ref{sec:proposedmethod}, fault diagnosis under different working conditions can be considered as separate tasks, with each task representing a multi-class classification problem. Table \ref{tab:cwru_tasks} provides detailed descriptions of each task, along with the relationship between motor load and motor speed. For each iteration, one of the tasks is treated as the target task, while the remaining tasks are used as auxiliary tasks. Following~\cite{wang2021footprint}, we chronologically divide the dataset of the target task into training, validation, and test sets with a ratio of 8:1:1.

% \subsubsection{Hyper-parameter Settings} 
%  The optimal hyperparameter settings for all models are empirically determined based on their performance on the validation set. Specifically, for TL-based models, the learning rate is selected from \{0.1, 0.05, 0.01, 0.005, 0.001, 0.0001\}, and the batch size is chosen from \{256, 128, 64\}. For meta-learning-based models, the learning rates $\alpha$ and $\beta$ are searched from \{0.1, 0.01, 0.001, 0.0001\}, while the batch size is again selected from \{128, 64, 32\} to ensure a fair comparison. For RT-ACM, the number of frozen layers, denoted as $l$, is explored within the range of [1,4], with a step size of one. The best setting, which is 3 layers, is found to be optimal for all tasks.

\subsubsection{Performance Comparison} 
\begin{figure}[tp]
    \centering
    \subfloat[]{\includegraphics[width=0.48\columnwidth]{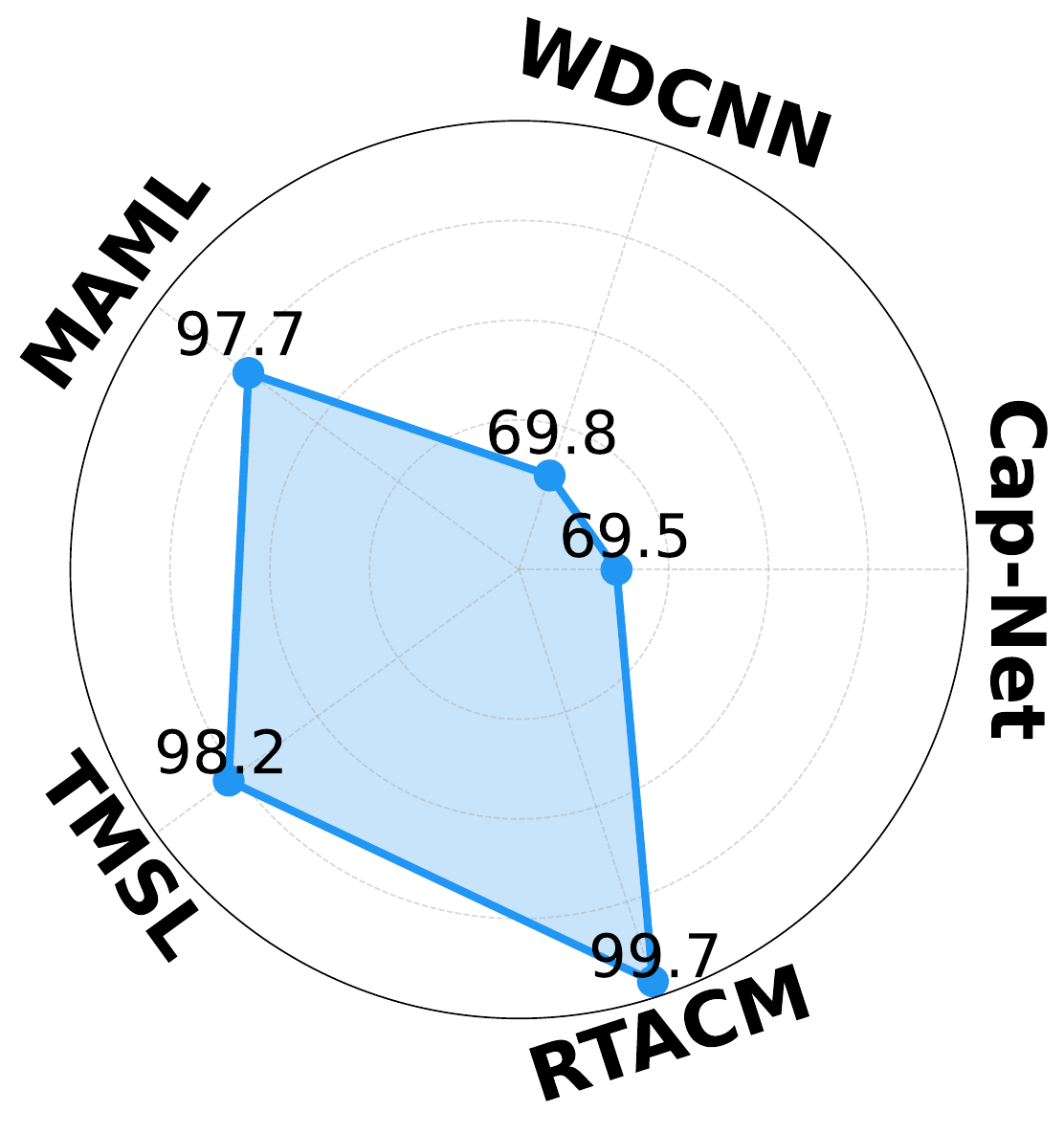}\label{Fig:radar_3}}
    \hfill
    \subfloat[]{\raisebox{-0.15em}{\includegraphics[width=0.48\columnwidth]{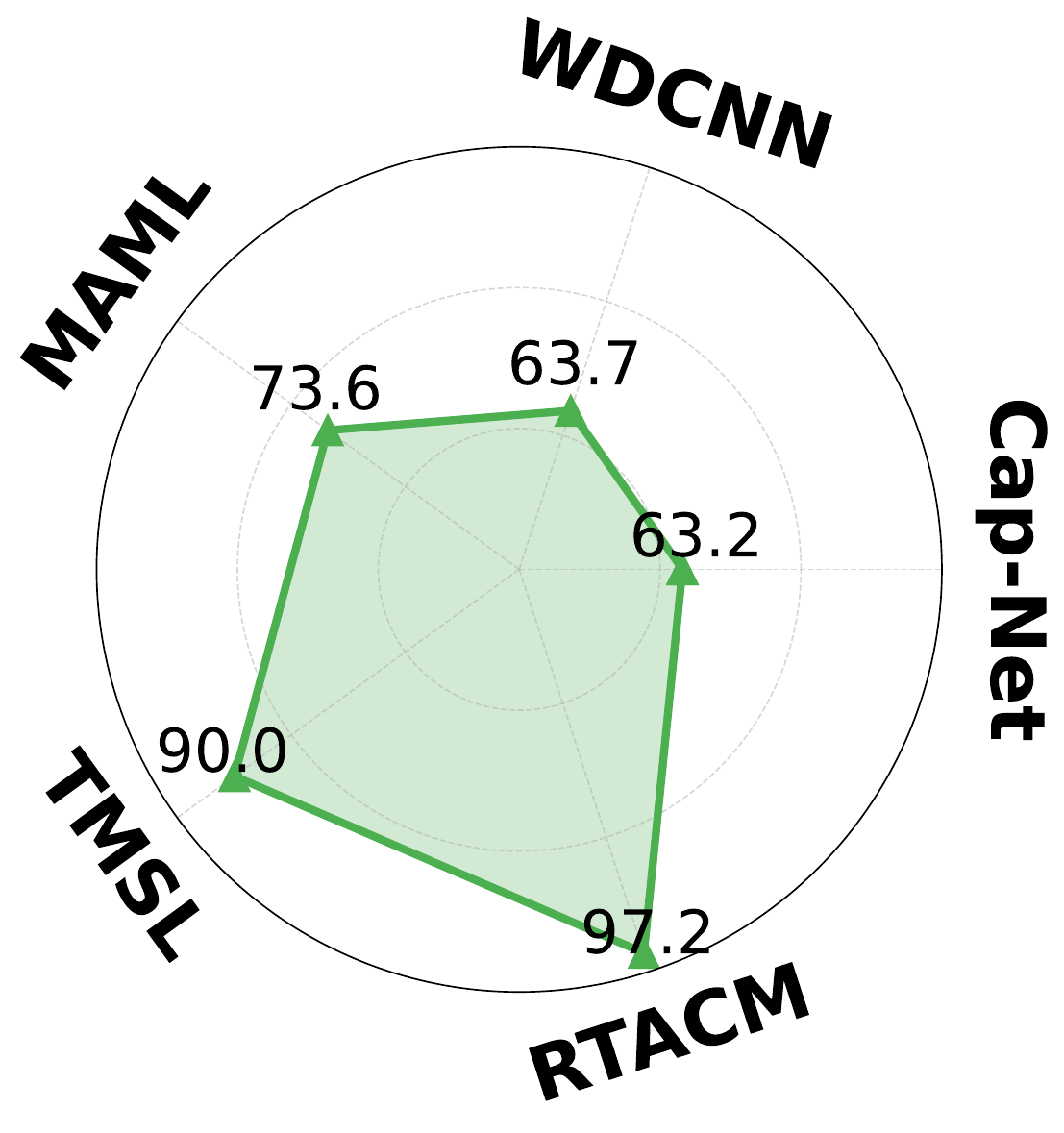}}\label{Fig:radar_6}}
    \caption{Comparison of (a) 3-Way and (b) 6-Way 5-Shot Performance on $T_1$.}
\end{figure}

Table \ref{tab:results_CWRU} presents a comprehensive comparison of deep learning-based models and meta-learning-based models across various tasks and shots. The results indicate a clear performance superiority of meta-learning-based models over their traditional counterparts. Specifically, our method RT-ACM consistently achieves the highest accuracy across all scenarios, notably outperforming other models with an impressive accuracy range of 98.51\% to 99.73\% in the 3-way tasks and 93.22\% to 97.48\% in the 6-way tasks. MAML and TMSL, while also meta-learning-based, demonstrate strong performances, especially in the 3-way 5-shot settings. Fig.~\ref{Fig:radar_3} presents the 3-way 5-shot performance results on Task 1, while Fig.~\ref{Fig:radar_6} illustrates the 6-way 5-shot performance on the same target task. It is evident that TMSL generally outperforms MAML, attributed to its advantage in sequence learning processes. Our proposed method RT-ACM integrates related task-aware meta-learning with a curriculum strategy, which further amplifies its performance. The related task-aware mechanism ensures that the model is adept at identifying and leveraging similarities between tasks, thereby improving its efficiency in learning new tasks. Additionally, the curriculum strategy within RT-ACM helps in systematically increasing the difficulty of the tasks presented during training, allowing the model to develop a deeper and more nuanced understanding of the task space. This dual approach not only enhances the learning process but also ensures that the model remains adaptable and highly accurate, as evidenced by its superior performance across all evaluated metrics.

\begin{figure}[tp]
    \centering
    \subfloat[]{\includegraphics[width=0.5\columnwidth]{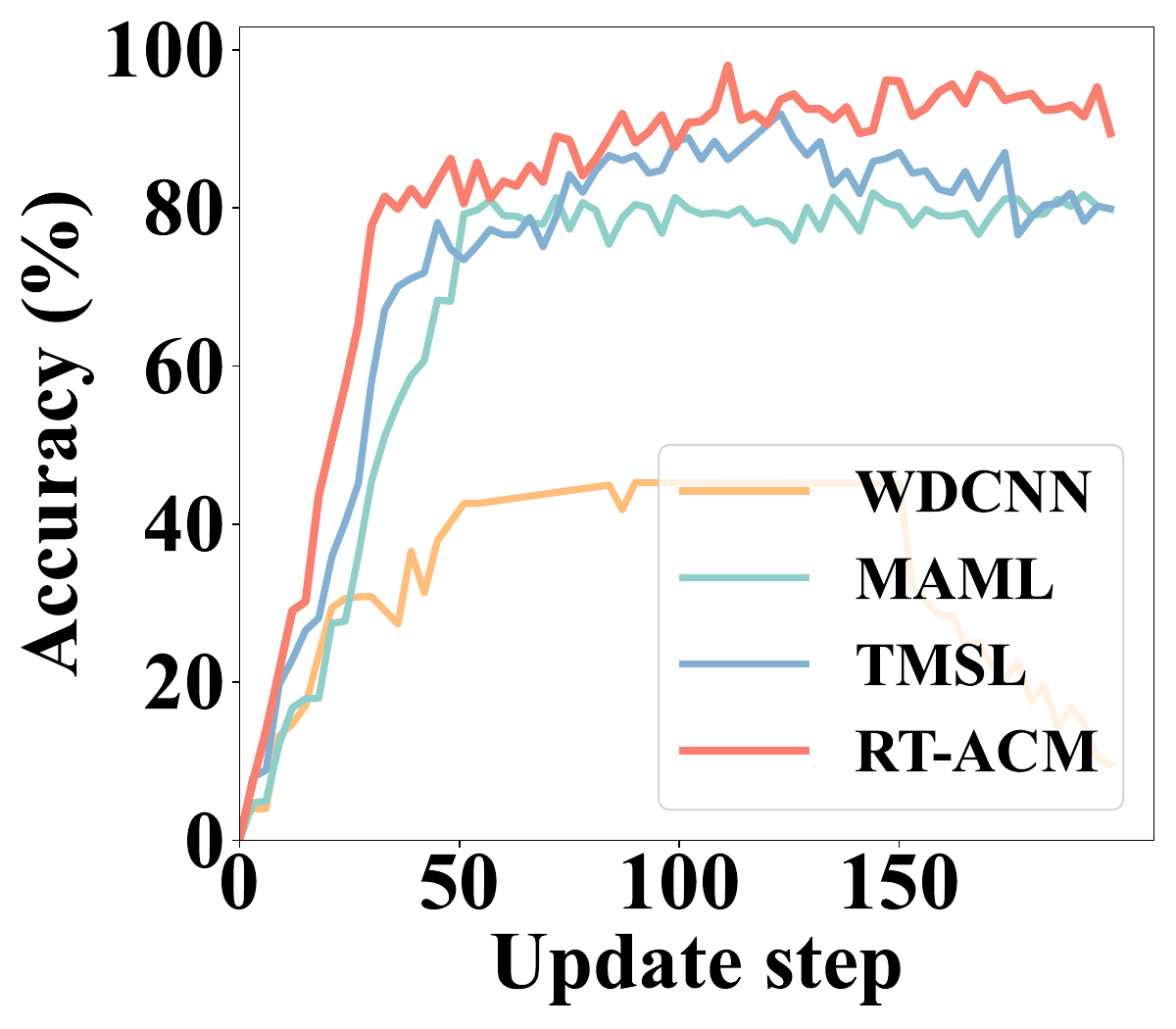}\label{Fig:Outter}}
    \hfill
    \subfloat[]{\raisebox{-0.15em}{\includegraphics[width=0.50\columnwidth]{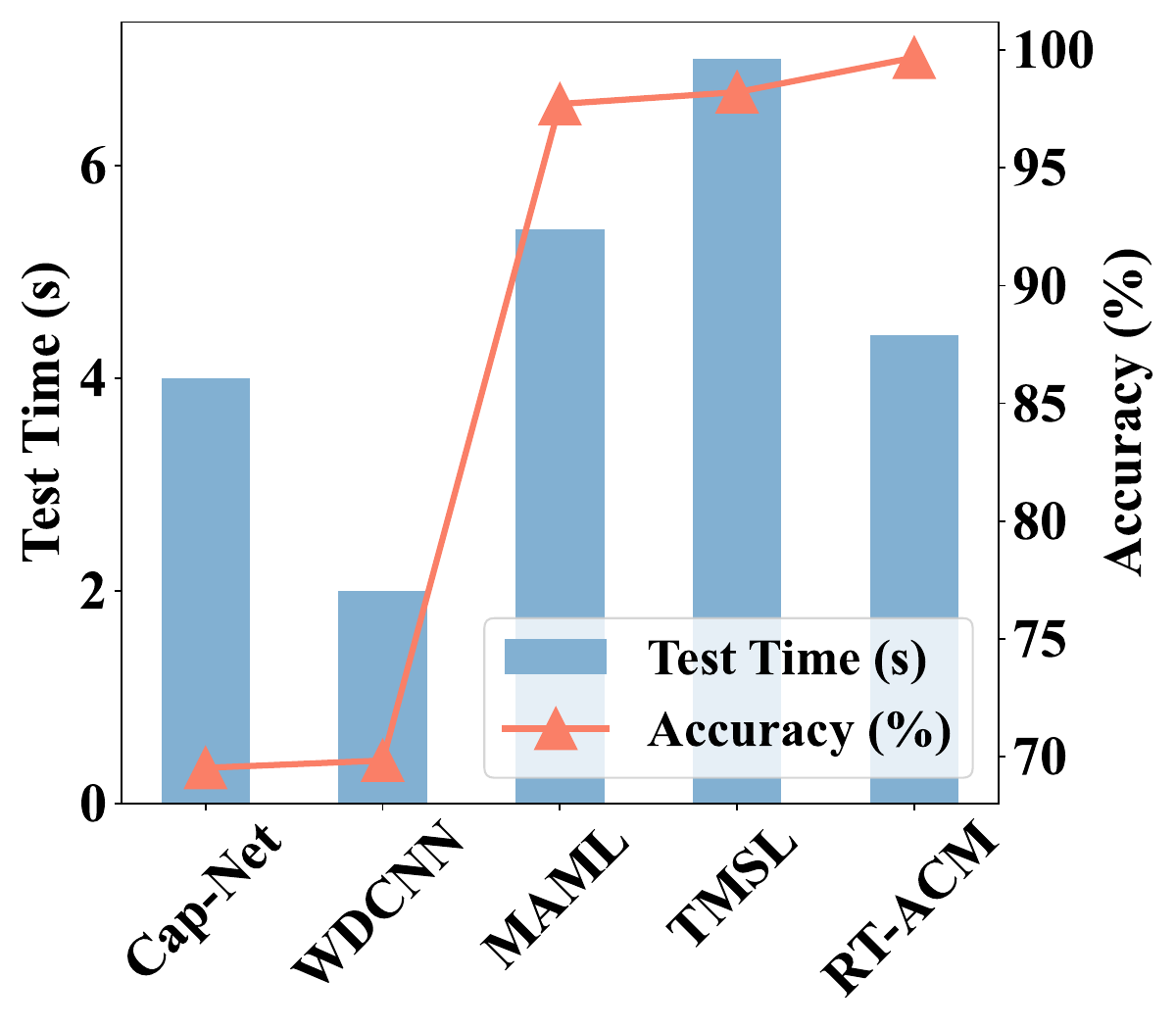}}\label{Fig:model_performance}}
    \caption{Accuracy in the (a) outer loop and (b) model performance on $T^1$.}
\end{figure}

Fig.~\ref{Fig:Outter} shows the performance comparison of four meta-learning-based models, in terms of accuracy over increasing global update steps. Our proposed method, RT-ACM, which leverages related task-aware meta-learning and a curriculum strategy, demonstrates a superior ability to rapidly achieve high accuracy. This rapid improvement can be attributed to the efficient transfer of knowledge from related tasks and a structured learning process that progresses from simpler to more complex tasks. In contrast, TMSL, which employs k-nn based task evaluation, lags behind RT-ACM and fails to match its performance throughout the training process due to a less effective related task prioritization. MAML, while maintaining steady progress, does not achieve the same level of accuracy as RT-ACM. WDCNN initially shows competitive accuracy but experiences a decline in performance in next stages, possibly due to overfitting or poor generalization. 

Fig.~\ref{Fig:model_performance} demonstrates the superior performance of the RT-ACM model in both running time in each iteration and accuracy on the 3-way 5-shot task. While RT-ACM requires slightly more computational time compared to deep learning models like Cap-net and WDCNN due to its meta-learning process of acquiring learning capabilities through different tasks, it significantly reduces the time consumption compared to other meta-learning approaches like MAML and TMSL. This improvement in efficiency can be attributed to the curriculum strategy's progressive learning approach, moving from simple to complex tasks. Furthermore, RT-ACM achieves competitive accuracy levels, highlighting its potential in balancing the trade-off between computational efficiency and model performance. The incorporation of auxiliary tasks has proven beneficial in improving accuracy on the target task.

\subsection{Case B: PU Bearing Dataset}
\subsubsection{Dataset Description} 
Different from CWRU dataset, the Paderborn University (PU) Bearing Dataset is a pivotal diagnostic tool for rolling bearings include the artificially damaged data. To mitigate the influence of similar bearings on the dataset, vibration signals were collected from 26 bearings characterized by artificial damage, natural damage, and healthy states. The detailed working conditions are presented in Table~\ref{tab:info_PU}. Following~\cite{hu2021task}, 32 unique bearing fault types are evaluated across four distinct operating conditions. To reflect the complexity of real-world industrial environments and to enhance the precision of diagnostic outcomes, each fault type is considered a separate category under each working condition, resulting in a total of 128 fine-grained fault categories. From these, 100 categories are randomly chosen for training purposes, while the remaining 28 are reserved for testing. Further specifics of the working conditions are provided in Table~\ref{tab:PU_tasks}.

\begin{table}[t]
\centering
\caption{THE COMPARATIVE METHODS IN PU DATASET  ($T^1$) \\(*The best results are highlighted in bold; the runner up is underlined.)}
\label{tab:PU_results}
\renewcommand{\arraystretch}{1.2}
\begin{tabular}{ccccccc}
\toprule
\hline
\multirow{3}{*}{\textbf{Models}} & \multicolumn{2}{c}{\textbf{Accuracy}} & \multicolumn{2}{c}{\textbf{Precision}} & \multicolumn{2}{c}{\textbf{F1-score}} \\
                                 & \multicolumn{6}{c}{\textbf{5-way}}                                                                                     \\
                                 & \textbf{5-shot}   & \textbf{6-shot}   & \textbf{5-shot}    & \textbf{6shot}    & \textbf{5-shot}   & \textbf{6-shot}   \\ \hline
Cap-Net                          & 46.61             & 48.85             & 46.52              & 48.03             & 46.36             & 48.47             \\
WDCNN                            & 72.48             & 74.06             & 71.46              & 74.57             & 68.47             & 75.08             \\
MAML                             & 89.65             & 90.9              & 87.90              & 88.92             & 89.06             & 90.77             \\
TMSL                             & {\ul 90.97}       & {\ul 91.05}       & {\ul 89.26}        & {\ul 90.23}       & {\ul 90.77}       & {\ul 90.95}       \\
\textbf{RTACM}                   & \textbf{94.34}    & \textbf{95.65}    & \textbf{94.21}     & \textbf{95.42}    & \textbf{94.27}    & \textbf{96.18}    \\ \hline
\bottomrule
\end{tabular}
\end{table}

\begin{table}[thp]
\caption{Information of Bearings Used from PU Dataset}
\label{tab:info_PU}
\centering
\begin{tabular}{cccc}
\toprule
\toprule
\textbf{No.} & \textbf{Bearing Code} & \textbf{Fault Type} & \textbf{Damage Method} \\ 
\midrule
1  & K003  & Healthy      & -                 \\
2  & K004  & Healthy      & -                 \\
3  & K005  & Healthy      & -                 \\
4  & K006  & Healthy      & -                 \\
5  & KA01  & Outer Race   & Multiple Damage   \\
6  & KA03  & Outer Race   & Electric Engraver \\
7  & KA05  & Outer Race   & Electric Engraver \\
8  & KA07  & Outer Race   & Drilling          \\
9  & KA08  & Outer Race   & Drilling          \\
10 & KA09  & Outer Race   & Drilling          \\
11 & KI01  & Inner Race   & Multiple Damage   \\
12 & KI03  & Inner Race   & Electric Engraver \\
13 & KI05  & Inner Race   & Electric Engraver \\
14 & KI07  & Inner Race   & Electric Engraver \\
15 & KI08  & Inner Race   & Electric Engraver \\
16 & KA04  & Outer Race   & Fatigue           \\
17 & KA15  & Outer Race   & Plastic Deform    \\
18 & KA16  & Outer Race   & Fatigue           \\
19 & KA22  & Outer Race   & Fatigue           \\
20 & KA30  & Outer Race   & Plastic Deform    \\
21 & KB23  & Inner/Outer  & Fatigue           \\
22 & KB27  & Inner/Outer  & Plastic Deform    \\
23 & KI04  & Inner Race   & Fatigue           \\
24 & KI16  & Inner Race   & Fatigue           \\
25 & KI17  & Inner Race   & Fatigue           \\ \hline
26 & KI21  & Inner Race   & Fatigue           \\
27 & K001  & Healthy      & -                 \\
28 & K002  & Healthy      & -                 \\
29 & KA30  & Outer Race   & Plastic Deform    \\
30 & KB24  & Inner/Outer  & Fatigue           \\
31 & KI14  & Inner Race   & Fatigue           \\
32 & KI18  & Inner Race   & Fatigue           \\
\bottomrule
\bottomrule
\end{tabular}
\end{table}

\begin{table}[tp]
\caption{DETAILED DESCRIPTIONS OF EACH TASK UNDER PU DATASET}
\label{tab:PU_tasks}
\begin{tabular}{cccc}
\toprule
\toprule
Task & Rational speed (rpm) & Load torque (Nm) & Radial force (N) \\ \toprule
$T^1$  & 1500                 & 0.7              & 400              \\
$T^2$  & 900                  & 0.7              & 1000             \\
$T^3$  & 1500                 & 0.1              & 1000             \\
$T^4$  & 1500                 & 0.7              & 1000             \\ \bottomrule
\bottomrule
\end{tabular}
\end{table}
\subsubsection{Performance Comparison}
\begin{figure*}[tp]
	\centering
	\subfloat[Cap-Net]{\includegraphics[width=.48\columnwidth]{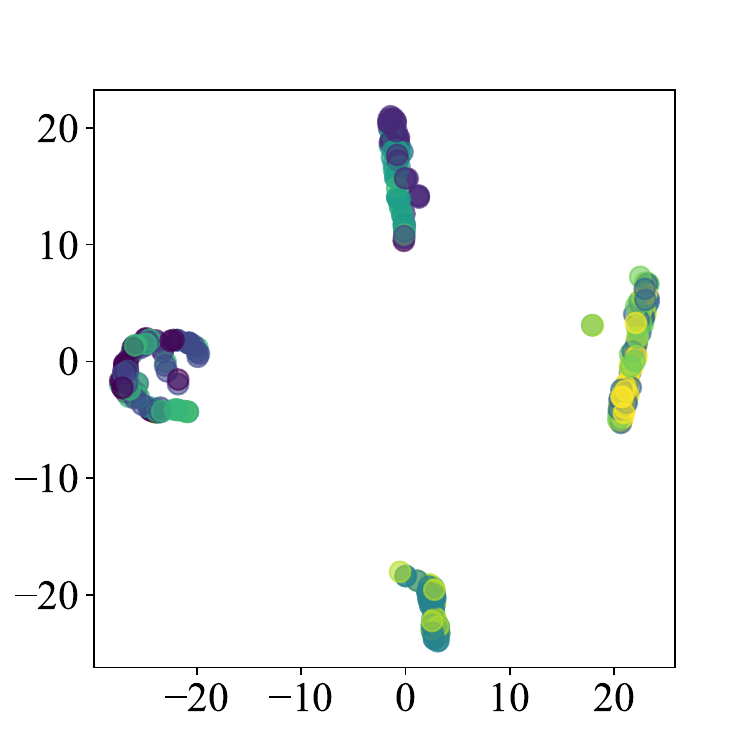}}\hspace{5pt}
	\subfloat[MAML]{\includegraphics[width=.48\columnwidth]{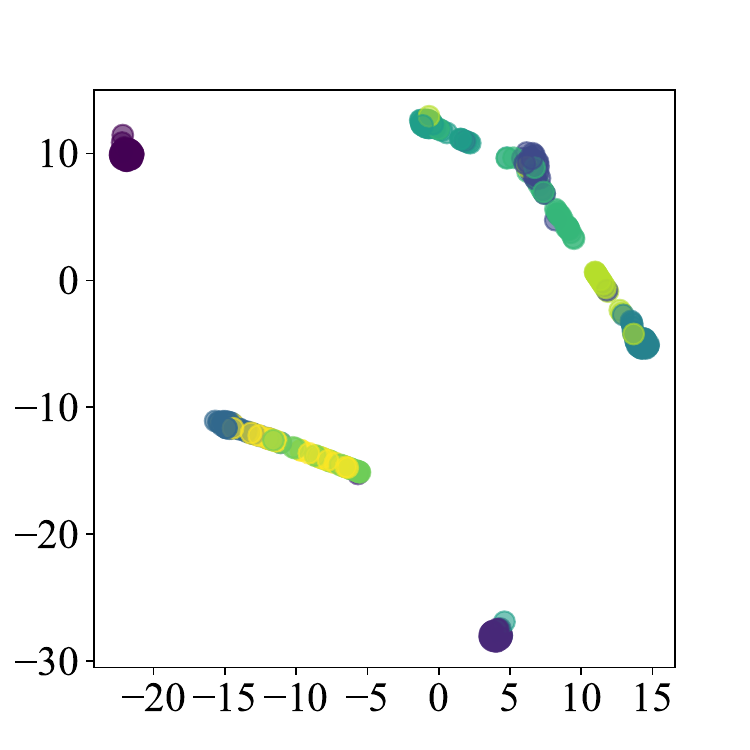}}
	\subfloat[TMSL]{\includegraphics[width=.48\columnwidth]{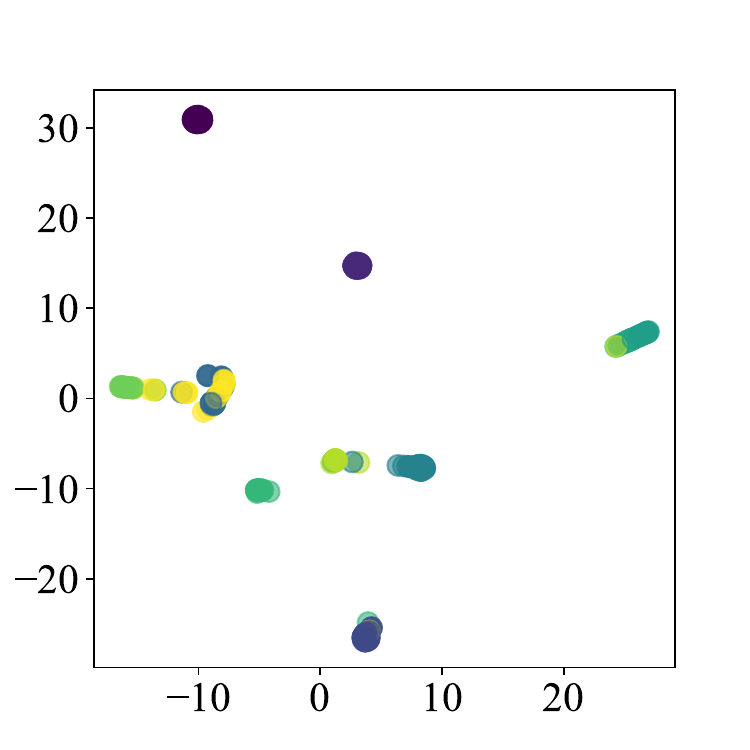}}\hspace{5pt}
	\subfloat[RT-ACM]{\includegraphics[width=.48\columnwidth]{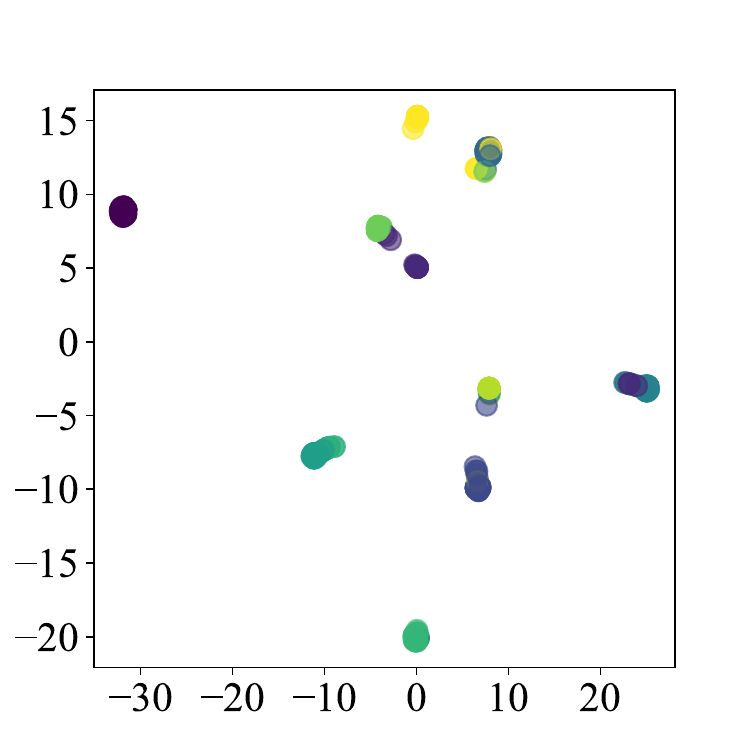}}
	\caption{The visualization of classification from four models on (10-way 5-shot).}
        \label{fig:classification}
\end{figure*}

Table~\ref{tab:PU_results} shows a performance comparison of various models on the PU dataset in 5-way, 5-shot and 6-shot. Notably, all methods except Cap-Net and WDCNN are meta-learning based approaches, allowing them to handle data sparsity issues more effectively than traditional DL-based model. Among these methods, MAML employs a meta-training process to integrate different tasks, enhancing its performance. TMSL utilizes a k-nn based sequence sampling in meta-training, providing a structured approach to learning. However, both MAML and TMSL do not adequately consider the diversity and relevance of auxiliary tasks in relation to the target task, which can limit their effectiveness. In contrast, our proposed method, RT-ACM, incorporates related task-aware meta-learning along with a curriculum strategy, resulting in superior performance. This is evidenced by RT-ACM achieving the highest scores across all metrics and shots, with an accuracy of 94.34\% and 95.65\% in 5-shot and 6-shot settings, respectively, outperforming the runner-up TMSL and other models significantly. Fig.~\ref{fig:classification} presents t-SNE visualizations for four models, highlighting their ability to discriminate between classes. The Cap-Net visualization shows poorly separated clusters, indicating significant class overlap and lower classification efficacy. MAML improves with tighter clusters but still suffers from overlap at the cluster centers. TMSL slightly betters class separation, with clearer clusters but some outliers. RT-ACM's visualization stands out, with each class forming distinct, isolated clusters, showcasing superior class discrimination and robust classification. The color-coded clusters denote various classes, and RT-ACM's close-knit clusters suggest strong internal similarity and external dissimilarity among classes, reflecting RT-ACM's effective feature space transformation for enhanced class distinction.

\begin{figure*}[tp]
	\centering
	\subfloat[Cap-Net]{\includegraphics[width=.48\columnwidth]{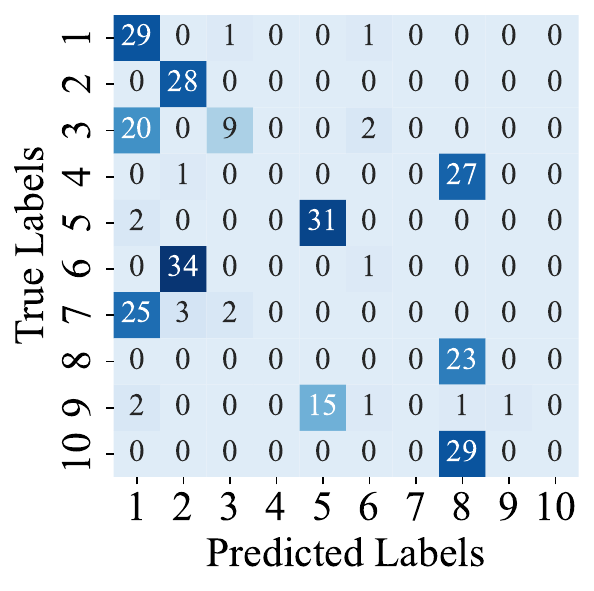}}\hspace{5pt}
	\subfloat[MAML]{\includegraphics[width=.48\columnwidth]{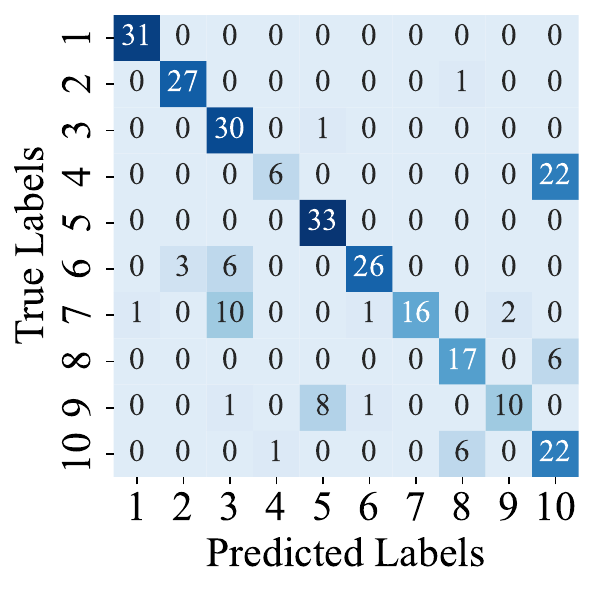}}
	\subfloat[TMSL]{\includegraphics[width=.48\columnwidth]{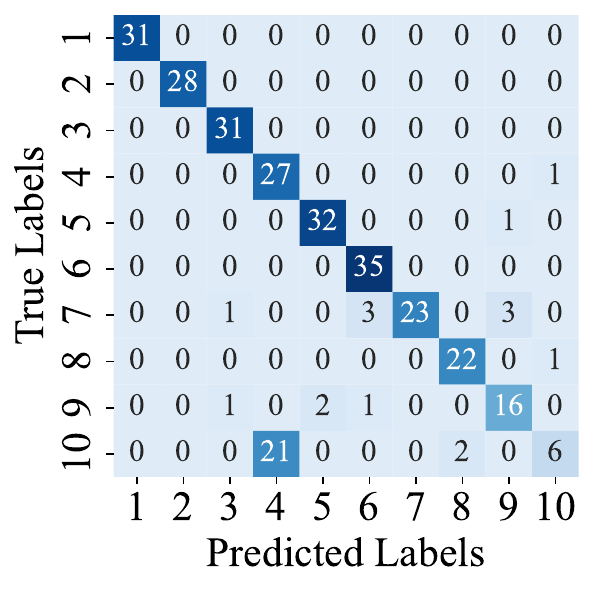}}\hspace{5pt}
	\subfloat[RT-ACM]{\includegraphics[width=.48\columnwidth]{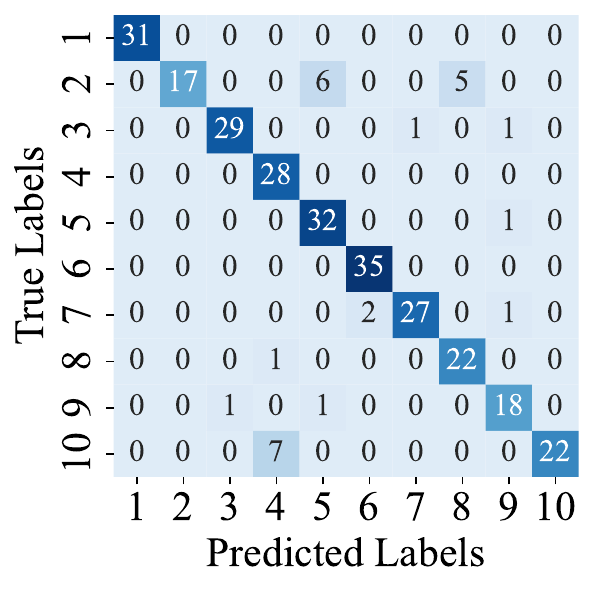}}
	\caption{The visualization of confusion matrices from four models (10-way 5-shot).}
        \label{fig:confusion_matrices}
\end{figure*}

Fig.~\ref{fig:confusion_matrices} showcases confusion matrices for four classification model, illustrating each model's predictive performance. The Cap-Net matrix shows widespread misclassifications, such as confusing class 3 with class 4, indicating difficulty in distinguishing similar classes. MAML improves, concentrating predictions along the diagonal but still mixes up classes like 4 and 9. TMSL further enhances accuracy, with clearer diagonal values and fewer errors, notably between classes 9 and 10. RT-ACM stands out with a strong diagonal focus, demonstrating superior accuracy and minimal misclassification, especially distinguishing between closely related classes like 4 and 10, where other models struggle. This highlights RT-ACM's robust predictive capability across diverse classes.

\begin{figure}[tp]
	\centering
	\subfloat[ ]{\includegraphics[width=.48\columnwidth]{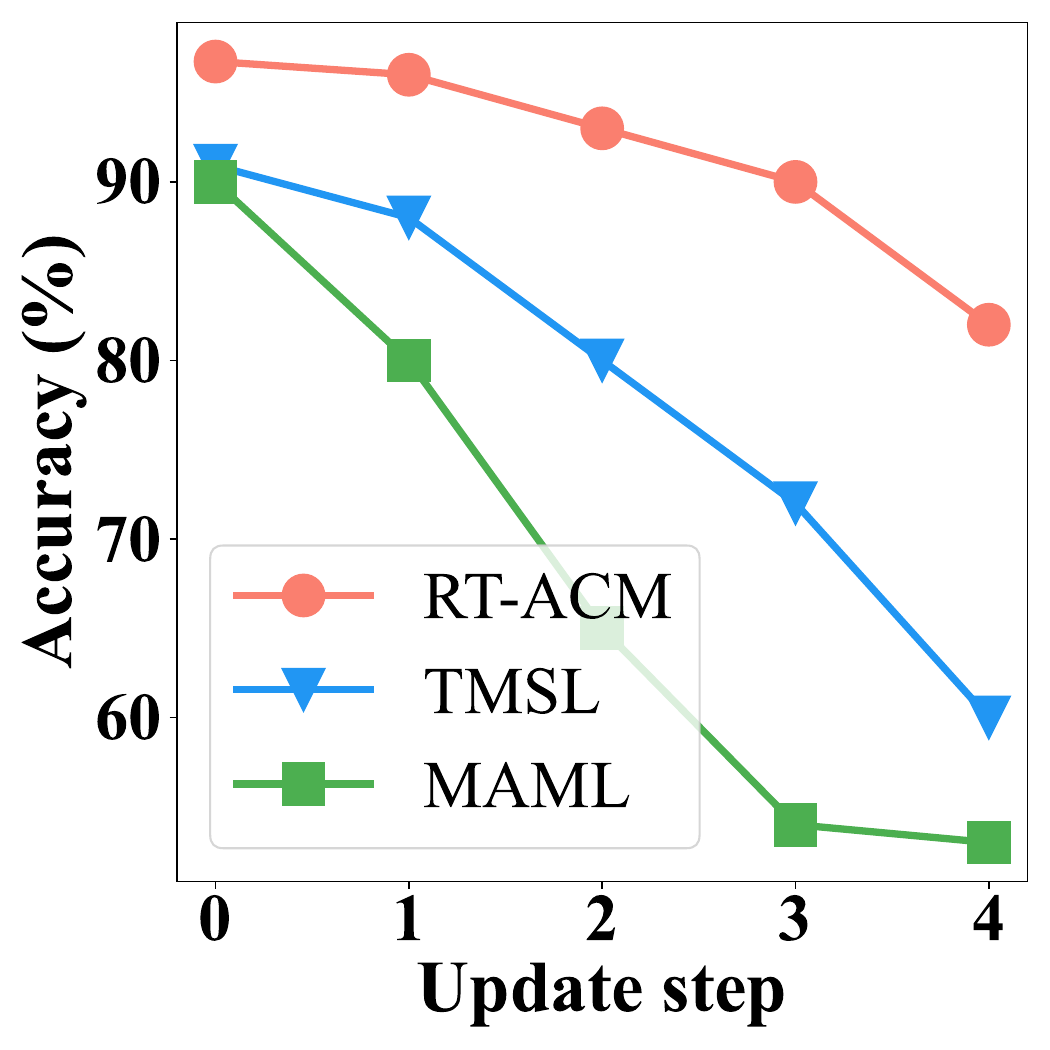}
    \label{fig:inner_acc}}
	\subfloat[ ]{\includegraphics[width=.49\columnwidth]{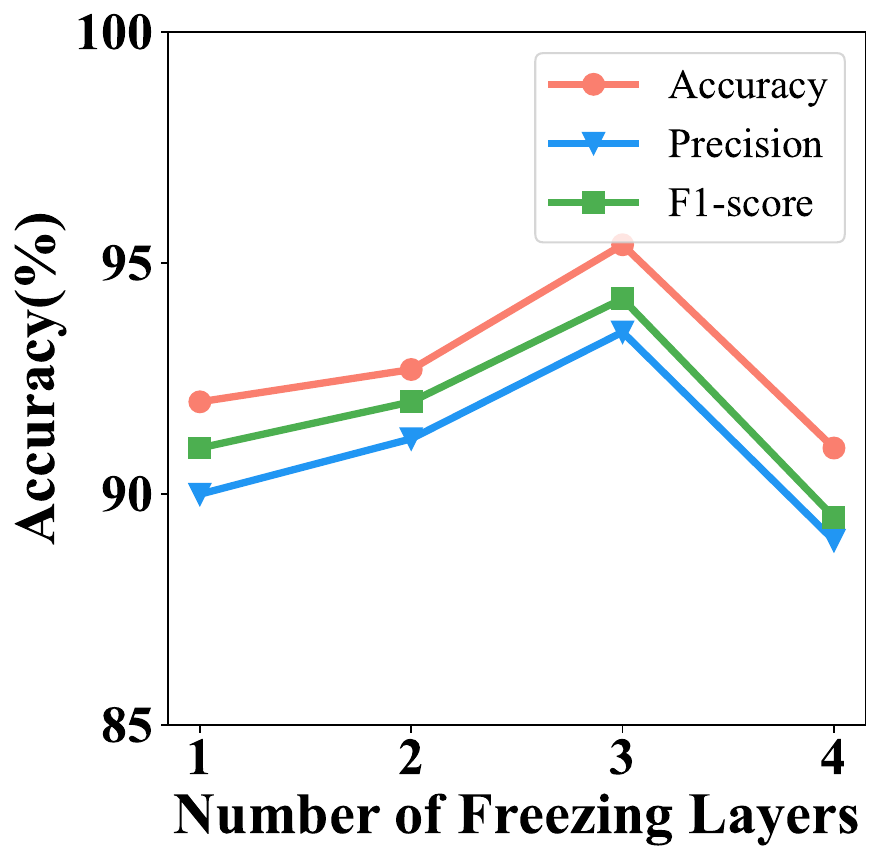}
    \label{fig:freeze_layer}}
	\caption{(a) The performance of models across varying numbers of local update steps and (b) the impact of the number of freezing layers.}
\end{figure}

Fig.~\ref{fig:inner_acc} illustrates a sensitivity analysis of a meta-learning-based model across varying numbers of local update steps. The results indicate that a single-step update is sufficient to significantly enhance accuracy, demonstrating that MAML-based models can achieve notable improvements with minimal updates. In Fig.~\ref{fig:freeze_layer}, the impact of the number of frozen layers on model performance is examined. The performance initially improves as the number of frozen layers increases, reaching a peak when three layers are frozen, after which performance slightly declines. This suggests that three frozen layers offer the optimal balance for performance. Too few frozen layers may lead to a loss of auxiliary information, while too many frozen layers limit the model’s ability to adapt to the target task, resulting in overfitting.

\section{Conclusion and Future Work}
\label{sec:conclusion}
In this paper, we introduced a Related Task Aware Curriculum Meta-learning (RT-ACM) enhanced fault diagnosis framework, a novel approach in the realm of fault diagnosis for smart manufacturing designed to address the issue of data scarcity. RT-ACM introduces an innovative approach by employing related task-aware meta-learning to enhance primary fault diagnosis tasks. It adheres to the principle of ``paying more attention on relevant knowledge” by prioritizing the most pertinent information and following an ``easy-to-hard” curriculum strategy. Our comprehensive evaluations across two real-world datasets confirm the effectiveness and superiority of the RT-ACM framework. The experiments not only highlight the efficacy of the framework but also validate the architectural choices, underscoring the potential of combining curriculum strategy with related task aware meta-learning for enhanced fault diagnosis. In future work, we will explore new strategies that can learn both temporal and spatial representations of auxiliary tasks when transferring knowledge to the target task.

\clearpage
\bibliographystyle{IEEEtran}
\bibliography{mybibliography}

\newpage

\end{document}